\UseRawInputEncoding

\documentclass[letterpaper, 10 pt, conference]{ieeeconf}  
\usepackage{amsfonts}
\usepackage{amsmath}
\usepackage{epsfig}
\usepackage{amssymb}
\usepackage{epsfig}
\usepackage{float}
\usepackage{graphicx}
\usepackage{color,hyperref}
\usepackage{multirow}
\usepackage{tipa}
\usepackage{booktabs} 
\usepackage{cite} 
\usepackage{bbding} 
\usepackage{soul}

\IEEEoverridecommandlockouts                              

\title{\LARGE \bf
Deep Virtual-to-Real Distillation for Pedestrian Crossing Prediction
}

\author{Jie Bai, Xin Fang, Jianwu Fang*, Jianru Xue, and Changwei Yuan
\thanks{Jie Bai is with School of Electronic and Control Engineering, Chang’an University, Xi’an 710064, China, and Xin Fang, Jianwu Fang (corresponding author) and Changwei Yuan are with College of Transportation Engineering, Chang’an University, Xi’an 710064, China. 
        {\tt\small fangjianwu@chd.edu.cn}}%
\thanks{Jianru Xue is with the Institute of Artificial Intelligence and Robotics, Xi’an Jiaotong University, Xi’an 710049, China.
        {\tt\small jrxue@mail.xjtu.edu.cn}}%
}

\begin{document}

\maketitle
\thispagestyle{empty}
\pagestyle{empty}

\begin{abstract}

Pedestrian crossing is one of the most typical behavior which conflicts with natural driving behavior of vehicles. Consequently, pedestrian crossing prediction is one of the primary task that influences the vehicle planning for safe driving. However, current methods that rely on the practically collected data in real driving scenes cannot depict and cover all kinds of scene condition in real traffic world. To this end, we formulate a deep virtual to real distillation framework by introducing the synthetic data that can be generated conveniently, and borrow the abundant information of pedestrian movement in synthetic videos for the pedestrian crossing prediction in real data with a simple and lightweight implementation. In order to verify this framework, we construct a benchmark with 4667 virtual videos owning about 745k frames (called Virtual-PedCross-4667), and evaluate the proposed method on two challenging datasets collected in real driving situations, i.e.,  JAAD and PIE datasets. State-of-the-art performance of this framework is demonstrated by exhaustive experiment analysis. The dataset and code can be downloaded from the website \footnote{\url{http://www.lotvs.net/code_data/}}.

\end{abstract}

\section{INTRODUCTION}
Vulnerable road users (pedestrians, cyclists and motorbikes) act the main role to occupy the right of way with vehicles, and take more than half of all road traffic deaths \cite{trafficdeath} investigated by World Heath Organization (WHO). The movement of them inevitably conflicts with the natural driving behavior of vehicles, and the crossing behavior is the most typical one. Facing the developing trend of autonomous driving or assisted driving systems, prediction of the crossing behavior of vulnerable road users is essential for safe driving.
 
\begin{figure}[!t]
  \centering
 \includegraphics[width=\hsize]{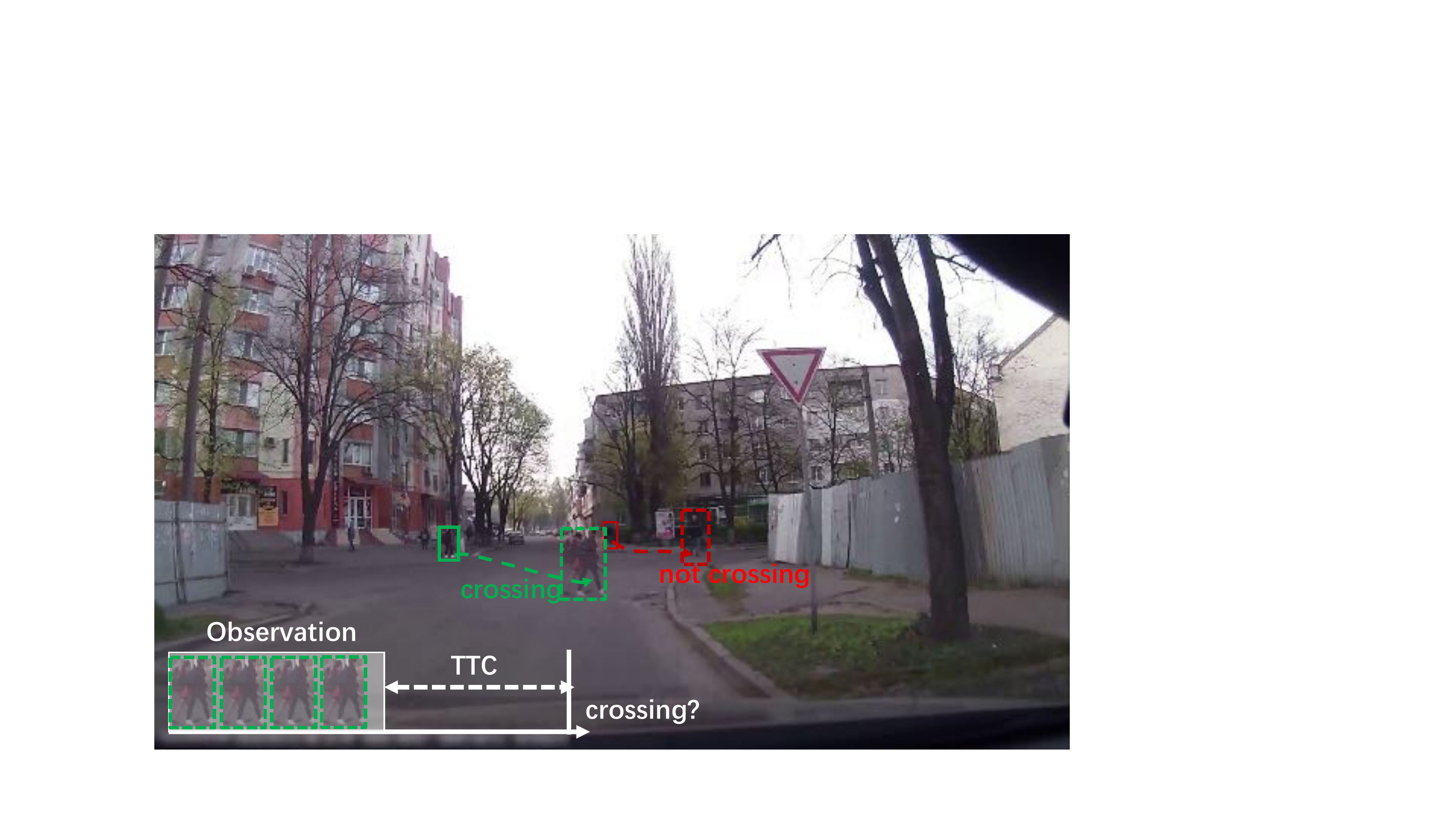}
  \caption{An example taken from Joint Attention for Autonomous Driving (JAAD) dataset \cite{DBLP:conf/iccvw/RasouliKT17} for the illustration the pedestrian crossing prediction, where TTC represents the time to cross interval, and the dash boxes represent the future location to be predicted.}
  \label{fig1}
  \vspace{-1em}
\end{figure}

In this work, we focus on the pedestrian crossing prediction problem that takes a historical video observation as input and predict whether the pedestrians cross or not in future time, demonstrated in Fig. \ref{fig1}. In this field, 2D poses \cite{DBLP:conf/iros/WangP20,DBLP:conf/itsc/CadenaYQW19}, pedestrian bounding boxes \cite{DBLP:journals/corr/abs-2107-08031}, optical flow \cite{DBLP:conf/cvpr/CarreiraZ17}, scene context \cite{DBLP:conf/bmvc/RasouliKT19}, vehicles speeds \cite{DBLP:journals/corr/abs-2104-05485}, trajectories \cite{DBLP:conf/itsc/XueLCZHZ19}, ego-motion of vehicles \cite{DBLP:journals/corr/abs-2104-05485} are utilized in previous works. In the meantime, the deep learning models, such as I3D \cite{DBLP:conf/cvpr/CarreiraZ17}, LSTM/RNN-based temporal models \cite{DBLP:conf/ivs/LorenzoPWSLS20,DBLP:conf/itsc/XueLCZHZ19}, as well as the transformers \cite{DBLP:journals/sensors/0002AIBHLS21} are adopted in recent years. However, because of the high-mobility of pedestrian, the prediction results of previous works do not approve each other \cite{DBLP:conf/wacv/KotserubaRT21}, especially for the starting time when the pedestrians show a small scale. In addition, the large scale change,  various light conditions, diverse weather conditions, complex camera motion of vehicles, etc., form challenges for this topic. 

In actual driving situations, pedestrian crossing behavior often appears in the road intersection and the main road. Meanwhile, for safe driving, the higher speed the vehicle takes, the earlier the pedestrian crossing should be predicted, even with a small scale. In addition, weather condition is another key issue in this field, and severe weather conditions, such as rainy, fogy, and snowy days cause an unclear demonstration of front pedestrians. However, these situations appear rarely in driving experience. Consequently, it is difficult to collect large-scale dataset covering different light and weather conditions and may take large laborious annotation work. Therefore, the aforementioned issues cause a few-shot problem that restricts the performance of pedestrian crossing prediction \cite{DBLP:conf/wacv/KotserubaRT21}, and result in one main problem:

\emph{How to collect enough pedestrian crossing data covering all kinds of light, weather, and occasion conditions? }

However, it is difficult to tackle this problem in practical driving. A recent work \cite{DBLP:journals/corr/abs-2107-08031} explored the virtual data simulated by CARLA \cite{Dosovitskiy17}, which collected a large-scale action prediction dataset and defined the crossing situation. Then, the work took a fine-tuning module to transfer the distribution of synthetic data to the one of real data. However, in order to make the distribution be transferable, the work \cite{DBLP:journals/corr/abs-2107-08031} only considers the bounding boxes of pedestrians. Nevertheless, bounding boxes have no scene information, and any crossing behavior (not limited to pedestrian crossing) could be treated as pedestrian crossing (false alarm), shown by Fig. \ref{fig1}. In the meantime, vast previous methods \cite{DBLP:conf/wacv/KotserubaRT21,DBLP:conf/iros/WangP20} verified that other appearance information, motion information, scene information, etc. are useful. However, more input information will cost more computational cost for representing the knowledge of these information, while we need a lightweight implementation in practical use. Therefore, different from the fine-tuning mode, we explore the knowledge distillation (KD) for pedestrian crossing detection. 

In order to leverage the knowledge of crossing behavior in synthetic data with various scene conditions, we formulate a deep Virtual-to-Real distillation framework for Pedestrian Crossing Prediction (named as VR-PCP). The distillation framework contains a teacher PCP network to be trained with synthetic videos, and a lightweight student PCP model for further implementation in some practical platforms. Meanwhile, the distillation framework can absorb the abundant information of motion, location, scene context of pedestrian crossing behavior with the help of teacher PCP, and transfer them to the student PCP. 
For training teacher PCP, a new pedestrian crossing prediction benchmark with 4667 synthetic videos owning 745k frames (called \textbf{Virtual-PedCross-4667}, described in the experiment) is constructed.
Based on the exhaustive experiments on two challenging datasets, i.e., Joint Attention for Autonomous Driving (JAAD) \cite{DBLP:conf/iccvw/RasouliKT17} and Pedestrian Intention Estimation (PIE) \cite{DBLP:conf/iccv/RasouliKKT19}, the proposed method outperforms other state-of-the-art ones. 

\section{RELATED WORK}

\subsection{Action Anticipation in Videos}

Action anticipation in videos pursues an accurate action prediction exhibited by the objects in upcoming video streams. Different from the pedestrian crossing behaviors, the actions may occur in any occasions but existing action anticipation works involve pedestrian participants. 

Most of the previous works exploited the spatial-temporal consistency between the actions in partially observed and the actions in the whole videos. Therefore, some works explored order relations of the local spatial-temporal features of the partially observed video aligned on the global features of whole video  \cite{DBLP:journals/pami/KongTF20,DBLP:journals/ijcv/WuWHLL21}. For an accurate prediction of actions, the alignment order of the local features of partially observed videos is important. Therefore, many works \cite{DBLP:journals/tcsv/ChenLSZ21,DBLP:journals/pami/KongTF20}focused on the global feature learning. In order to obtain an accurate action anticipation, beside the spatial-temporal consistency consideration, the action semantic consistency is also important. For this information, many works extract the semantic features from object interaction \cite{DBLP:journals/tip/RoyF21,DBLP:conf/eccv/LiuTLR20}, structural or hierarchical graph relation of scene \cite{DBLP:conf/eccv/LanCS14,DBLP:journals/ijcv/WuWHLL21}, scene context information \cite{DBLP:conf/cvpr/KongTF17,DBLP:journals/tcsv/HuangLLLL20}, and so on.

As for our work, the most related kind of works are the action anticipation in the egocentric videos (also called first-person videos) \cite{DBLP:conf/ijcai/ZhangMYLJR21}. In this domain, the observers' intention, the interaction of the observer with the objects in the scene were concentrated. Liu \emph{et al.} \cite{DBLP:conf/eccv/LiuTLR20} explored the intentional hand movement, and jointly learned the deep relation of egocentric hand motion, interaction hotspots and future action.  Zhang \emph{et al.} \cite{DBLP:conf/ijcai/ZhangMYLJR21} proposed a counterfactual analysis framework to infer the semantic and visual causal features of actions. 
 
\subsection{Pedestrian Crossing Prediction}

In pedestrian crossing prediction, many works formulate the pedestrian crossing prediction problem as a pedestrian trajectory prediction task. For example, Xue \emph{et al.} \cite{DBLP:conf/itsc/XueLCZHZ19} proposed an encoder-decoder LSTM network to predict the trajectory of crossing pedestrians.  Wu \emph{et al.} \cite{DBLP:conf/itsc/WuWZXW20} proposed a pedestrian trajectory prediction method which involves the intention and behavior information of pedestrians in prediction. In addition, pedestrians to cross usually demonstrate an interaction with upcoming vehicles, i.e., with an intention communication. Commonly, the body pose and the eye-gaze direction are the two main signal in communication. Therefore, some works investigated the pedestrian pose \cite{DBLP:conf/ivs/LorenzoPWSLS20,DBLP:conf/iros/WangP20}, joint attention \cite{DBLP:conf/iccvw/RasouliKT17}, etc., to encode the crossing features. The famous work is the Joint Attention for Autonomous Driving (JAAD) \cite{DBLP:conf/iccvw/RasouliKT17}.

Pedestrian crossing has the special context information of occurrence, i.e., marked by the road boundaries. Hence, context information was exploited widely in the pedestrian crossing prediction task \cite{DBLP:conf/ijcai/YaoAJV021,DBLP:conf/iros/SchneemannH16}. For instances, Rasouli \emph{et al.} \cite{DBLP:conf/bmvc/RasouliKT19} took the scene dynamics and visual feature of the pedestrians into account, and proposed a stacked RNN to infer the temporal prediction. In the same research group, they constructed the famous Pedestrian Intention Estimation (PIE) dataset \cite{DBLP:conf/iccv/RasouliKKT19}.

Because of the rarity of the pedestrian crossing in unmarked road, Achaji \emph{et al.} \cite{DBLP:journals/corr/abs-2107-08031} proposed a new work trained on the large-scale simulated data, and it emphasis that using only bounding boxes of pedestrian can leverage an accurate pedestrian crossing prediction. However, this work does not consider the few-shot problem of the pedestrian crossing problem when encountering severe weather condition, low light conditions.  In addition, pedestrian crossing stands on the special context of road region, while only the bounding box information cannot reflect road context, and any movement similar to the crossing behavior could be accepted.

\subsection{Knowledge Distillation for Behavior Prediction}

Relating to this work, Knowledge Distillation (KD) recently is used in the action prediction task in general videos \cite{DBLP:conf/aaai/CaiLHZ19,DBLP:conf/icpr/CamporeseCFFB20}, which transfers the complete information in other full videos into the partially observed videos for the future action prediction.  For example, Camporese \emph{et al.} \cite{DBLP:conf/icpr/CamporeseCFFB20} used the knowledge distillation to smooth the labels between the full videos and the partial videos, and the teacher model extracted the semantic prior information for the action anticipation. Wang et al. \cite{DBLP:conf/cvpr/WangHLZZ19} proposed a teacher model for recognizing actions from full videos, and a student model for predicting early actions from partial videos. Similarly, the feature embeddings and action classifiers trained on the full videos are distilled in the teacher model for the prediction. Recently, collaborative knowledge distillation is designed to tackle the action prediction observed in multi-view cameras.

Nowadays, KD is absent in pedestrian crossing prediction. As for the few-shot problem or the sample shortage issue in the complex environment conditions, KD may be useful and will be investigated in our work.

\section{METHOD}

\subsection{Problem Formulation}
With the help of the synthetic data, we formulate a virtual-to-real distillation framework to boost the performance of pedestrian crossing prediction on the real data (\emph{abbrev.} VR-PCP). Consequently, this work builds a model with a teacher model $\mathcal{T}$ trained on the synthetic data, and a student model $\mathcal{S}$ adapting to the real data. Under this framework, two concerns are how to borrow the abundant features learned by $\mathcal{T}$ on the synthetic data to $\mathcal{S}$, and make $\mathcal{S}$ be easy and lightweight for implementation. 

 \begin{figure}[htpb]
  \centering
  \includegraphics[width=\hsize]{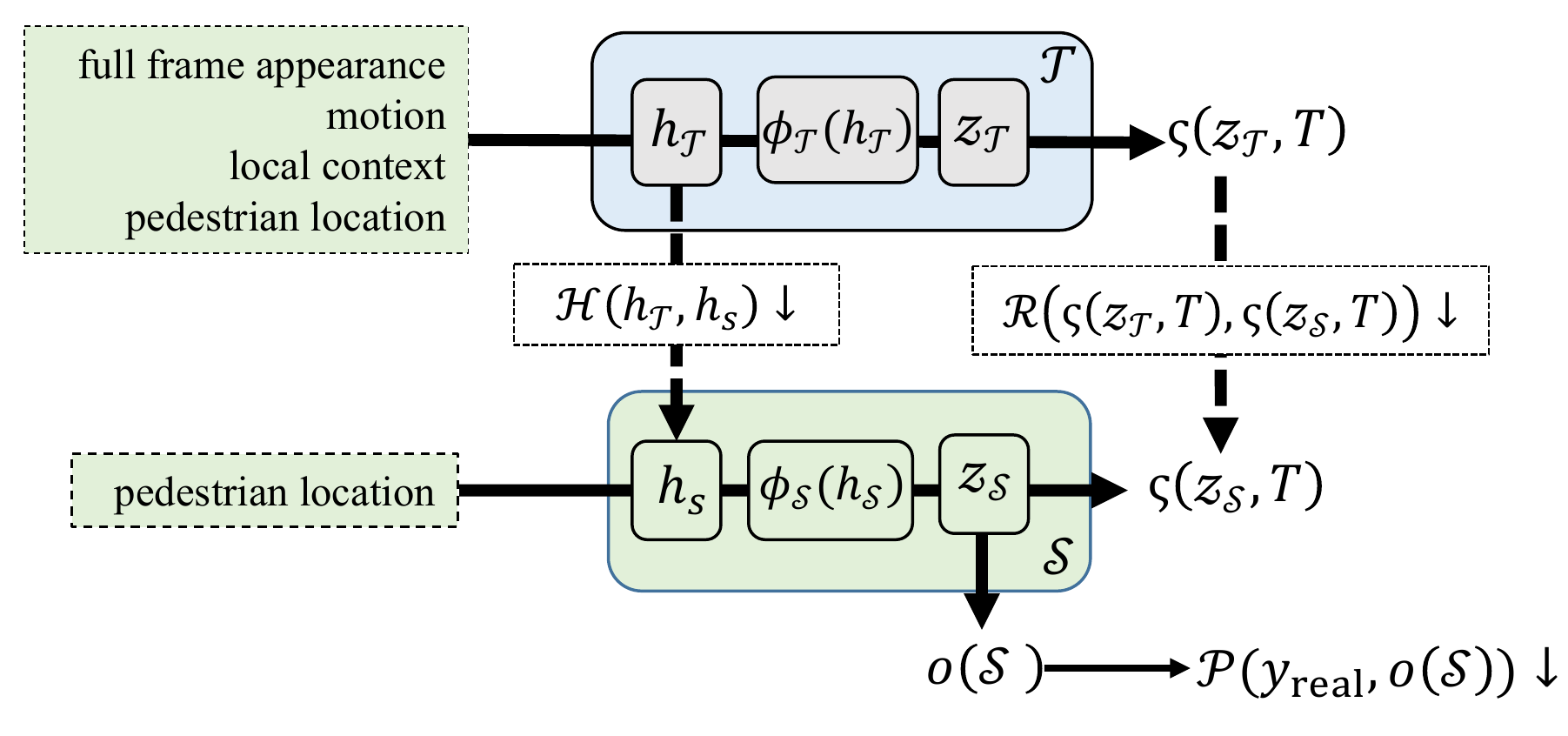}
  \caption{The overall formulation of VR-PCP.}
  \label{fig2}
\end{figure}
Therefore, we define the problem in this work as minimizing the following objective function.
\begin{equation}
\begin{array}{ll} 
\mathcal{L}=\mathcal{R}(\zeta(z_\mathcal{T},T),  \zeta(z_\mathcal{S},T))+\mathcal{H}(h_\mathcal{T},h_\mathcal{S})\\
\verb'   '+\mathcal{P}(y_{real},o(\mathcal{S})),\\
\end{array}
\label{eq1}
\end{equation}
where $\mathcal{R}(.)$ represents the response distillation function, which transfers the logits $z_\mathcal{T}$ outputted by the teacher model to the ones $z_\mathcal{S}$ of student model. Here the logits denote a 2-dimensional indicator of crossing or not-crossing. $\mathcal{H}(.)$ denotes the feature distillation function which absorbs the abundant feature information $h_\mathcal{T}$ learned by  $\mathcal{T}$ on synthetic data to the features $h_\mathcal{S}$ in training $\mathcal{S}$ on real data. $\mathcal{P}$ specifies the prediction model using $\mathcal{S}$ to approximate the  ground-truth ($y_{real}$) of crossing or not crossing in real data. $\zeta(.)$ and $o(\mathcal{S})$ are the probability function of logits distillation and the crossing prediction function of $\mathcal{S}$, respectively. $\zeta(.)$ is commonly defined as a softmax function with a hyper-parameter $T$ which controls the importance of each value of logits, and larger $T$ pursues a higher importance. For a clear demonstration of this formulation, we demonstrate the overview of VR-PCP in Fig. \ref{fig2}.

The logits $z_\mathcal{T}$ is the target vector (crossing or not-crossing) before normalization and is obtained by a transformation of features in the hints layer of the teacher model, and is denoted as $\phi_\mathcal{T}(h_\mathcal{T})$. Similarly, $z_\mathcal{S}$=$\phi_\mathcal{S}(h_\mathcal{S})$. In addition, the prediction model $o(\mathcal{S})$ is denoted as $\zeta(z_\mathcal{S},T=1)$, which denotes a universal softmax function. 
Consequently, the pedestrian crossing prediction problem can be formulated as modeling the teacher PCP model $\mathcal{T}$, the student PCP model $\mathcal{S}$, and the distillation functions of $\mathcal{R}(.)$ and $\mathcal{H}(.)$, as well as the prediction model $\mathcal{P}(.)$.

\subsection{Modeling $\mathcal{T}$}

 \begin{figure}[!t]
  \centering
  \includegraphics[width=\hsize]{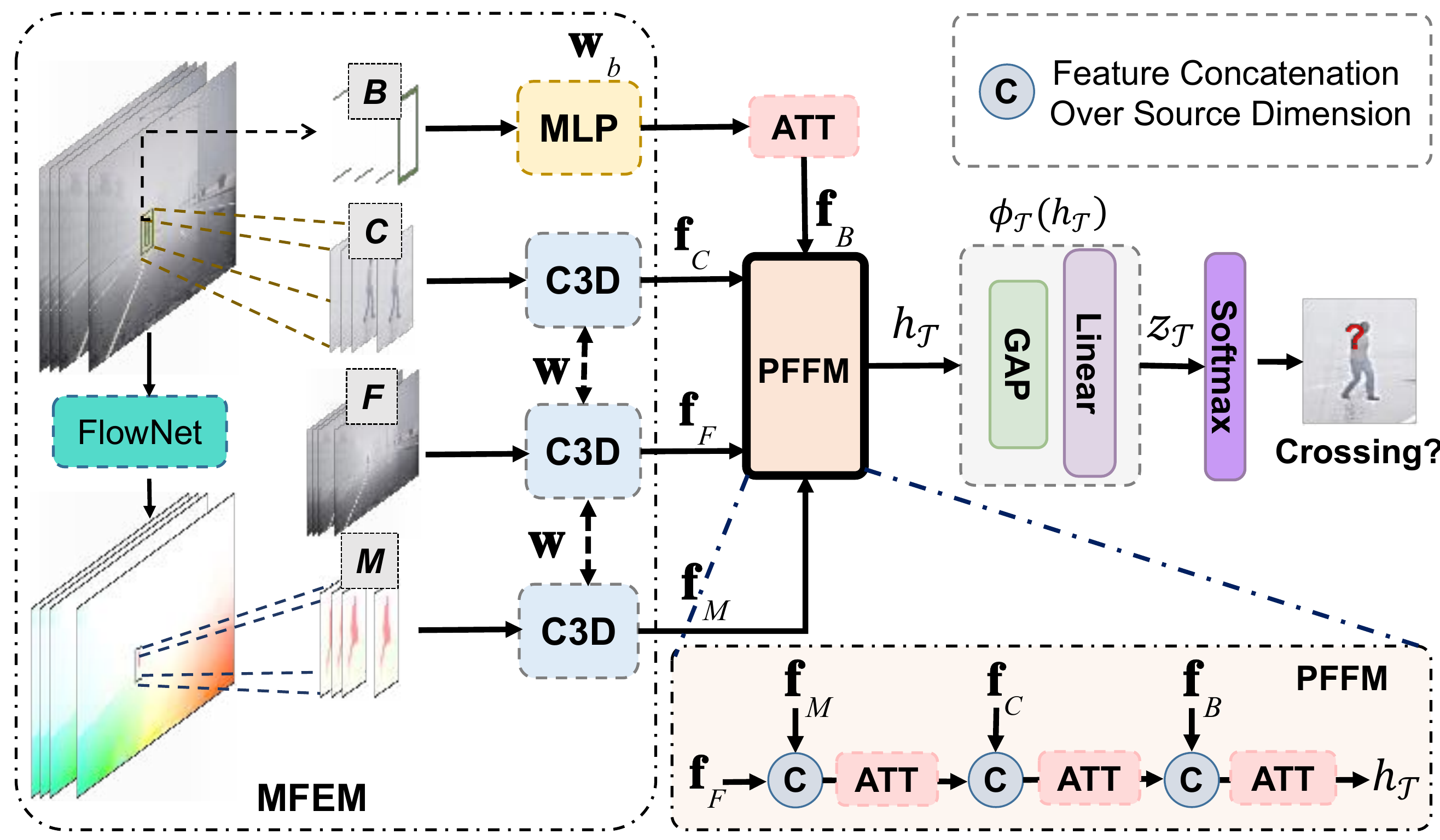}
  \caption{The overall network structure of the teacher PCP model $\mathcal{T}$.}
  \label{fig3}
\end{figure}

We all know that in the knowledge distillation framework, the performance of the student model relies on the learning ability of the teacher model. Commonly, teacher model owns a complex architecture for learning the rich knowledge in the dataset. In order to model the teacher PCP model $\mathcal{T}$ better, we model it with a cross-modal feature fusion module. Specifically, the bounding boxes, local context, motion of pedestrian, and the full frame are considered to fulfill a rich feature learning. Notably, we add the full frame into the feature learning to take account of the road structure. The schematic flowchart of $\mathcal{T}$ is shown in Fig. \ref{fig3}. We can see that the teacher PCP model has two parts of modules: the \emph{Multi-modal Feature Embedding Module} (MFEM) and a \emph{Progressive Feature Fusion Module} (PFFM). 

\subsubsection{MFEM}
Assume we input $N$ frames ${\bf{\emph{F}}}=\{f_t\}_{t=1}^N$, and predict the crossing or not-crossing label for the $f_m^{th}$ frame, where the time to crossing (TTC) interval is  $f_m-f_N$ frames. For motion information, we chose the optical flow image. Assume the local context region, bounding boxes, and the local motion region of one pedestrian are denoted as ${\bf{\emph{C}}}=\{{\bf{c}}_t\}_{t=1}^N$, ${\bf{\emph{B}}}=\{{\bf{b}}_t\}_{t=1}^N$, and ${\bf{\emph{M}}}=\{{\bf{m}}_t\}_{t=1}^{N-1}$, respectively. ${\bf{c}}_t$ is a sub-region of $f_t$ and it surrounds ${\bf{b}}_t$ with 1.5 times of the size of ${\bf{b}}_t$. Similarly, the scale of ${\bf{m}}_t$ is the same as the one of ${\bf{c}}_t$. The motion region ${\bf{m}}_t$ is actually the sub-region of optical flow image obtained by Flownet 2.0 \cite{DBLP:conf/cvpr/IlgMSKDB17} on frame $f_t$ and $f_{t-1}$.  In this work, we adopt 3 layers MLP for obtaining the embedding of each ${\bf{b}}_t$ and 3D convolution network (C3D) \cite{DBLP:conf/iccv/TranBFTP15} for the features of ${\bf{\emph{F}}}$, ${\bf{\emph{C}}}$, and ${\bf{\emph{M}}}$, as
\begin{equation}
\begin{array}{ll} 
{\bf{f}}_t=\text{MLP}({\bf{b}}_t, {\bf{w}}_b), 
{\bf{f}}_F=\text{C3D}({\bf{\emph{F}}}, {\bf{w}}),\\
{\bf{f}}_C=\text{C3D}({\bf{\emph{C}}}, {\bf{w}}),
{\bf{f}}_M=\text{C3D}({\bf{\emph{M}}}, {\bf{w}}),
\end{array}
\end{equation}
where the feature embedding of motion, local context region and full frame share the same weight ${\bf{w}}$, and ${\bf{w}}_b$ is the weight of 3-layers MLP. ${\bf{f}}_F$, ${\bf{f}}_C$, and ${\bf{f}}_M$ are with the same shape of 128-dimension and are the feature vectors of pedestrian location, frame appearance, local context and local motion region, respectively. 

\subsubsection{PFFM}
Although we have multiple kinds of information in the synthetic data, the importance of them may be different for pedestrian crossing prediction. Hence, this work designs a Progressive Feature Fusion Module (PFFM) for selecting the useful multi-modal features from global to local perspective, and each fusion step is fulfilled by a self-ATTention module (ATT).  PFFM emphasizes on the local feature more because of the closer relation to pedestrian and weaker disturbance of the background. 

For location feature vector ${\bf{f}}_t$ of pedestrian at time $t$, this work extracts the temporal attention feature ${\bf{f}}_B$ of locations over $N$ frames, and explores the temporal importance of time steps. Therefore, we also take the learnable self-ATTention module (ATT), and is defined as:
\begin{equation}
{\bf{f}}_B=\text{ATT}([{\bf{f}}_1;...;{\bf{f}}_t;...;{\bf{f}}_N]),
\end{equation}
where [;] denotes the stack concatenation over temporal dimension. 
Then, we fusion the location feature, frame feature, local context feature, and motion feature with a progressive fusion strategy, and specified as:
 \begin{equation}
h_{\mathcal{T}}=\text{ATT}([{\bf{f}}_B;\text{ATT}([{\bf{f}}_C;\text{ATT}([{\bf{f}}_M;{\bf{f}}_F])])]).
\end{equation}
Here, [;] represents the feature concatenation over feature modal dimension. The concatenated feature representation is defined as ${\bf{f}}_{fuse}\in R^{128\times K}$, where $K$ denotes the number of feature modals, and set as 2 for progressive fusion strategy. $\text{ATT}({\bf{f}}_{fuse})$ is defined as:
\begin{equation}
\begin{array}{ll} 
S=softmax({\bf{f}}_{last}W_{c1}[{\bf{f}}_{fuse}]),\\
\text{ATT}({\bf{f}}_{fuse})=tanh(W_{c2}[{\bf{f}}_{fuse}S;{\bf{f}}_{last}]),
\end{array}
\end{equation}
where ${\bf{f}}_{last}$ denotes the last column value of ${\bf{f}}_{fuse}$, and $W_{c1}$ and $W_{c2}$ are the weights of fully-connected layers. 

With $h_{\mathcal{T}}$, it is judged by a binary classifier $\phi(h_{\mathcal{T}})$ modeled by a Global Average Pooling (GAP) layer, 2 fully-connected layers, and generates the teacher logits $z_{\mathcal{T}}$ which is classified by a \emph{softmax} layer for crossing or not-crossing determination.

\subsection{Modeling $\mathcal{S}$}

With the rich information learning by $\mathcal{T}$, we aims to design a lightweight student PCD network $\mathcal{S}$ for practical implementation. In order to obtain the lightweight student PCD, there are two main considerations: 1) simplifying the input information, and 2) reducing the parameters of $\mathcal{S}$. In this work, we consider these two insights simultaneously.

For simplifying the input information, we only choose the pedestrian location ${\bf{\emph{B}}}=\{{\bf{b}}_t\}_{t=1}^N$ over $N$ frames, where ${\bf{b}}_t$ is a 4-dimensional information of ($x$, $y$, $height$, $width$), and ($x$,$y$) denotes the center coordinate. The pedestrian location information over $N$ frames is fed the student PCP model $\mathcal{S}$ implemented by some lightweight networks.

Although $\mathcal{S}$ pays more attention to use the location information of pedestrians, after the distillation of $\mathcal{T}$, it can absorb the information of motion, scene context, etc., which includes the influence of different lighting and weather conditions. Therefore, in real-world use, the distilled $\mathcal{S}$ can be effective and meaningful for pedestrian crossing task.

For reducing the parameters of $\mathcal{S}$, we take four kinds of lightweight architectures. They are naive \emph{Transformer} \cite{DBLP:conf/nips/VaswaniSPUJGKP17}, \emph{ResNet18} \cite{DBLP:conf/cvpr/HeZRS16}, \emph{MobileNet} \cite{DBLP:journals/corr/HowardZCKWWAA17} and \emph{ShuffleNet} \cite{DBLP:conf/cvpr/ZhangZLS18} with the weight size of 4.77M, 10.79M, 3.32M and 2.12M, respectively. In order to make the pedestrian location information be directly used in these lightweight networks, we introduce an embedding layer for transferring ${\bf{\emph{B}}}\in R^{N\times4}$ into ${\bf{\hat{\emph{B}}}}\in R^{N\times4\times 64}$ with the embedding of 64-dimension. The embedding layer is fulfilled by:
\begin{equation}
\begin{array}{ll} 
{\bf{\hat{\emph{B}}}={\bf{\emph{B}}}\odot{\bf{\hat{w}}}}+b
\end{array}
\end{equation}
where ${\bf{\hat{w}}}\in R^{4\times 64}$ is the weight of the embedding layer, $b$ is the bias value, and $\odot$ denotes the operation of dot product.
${\bf{\hat{\emph{B}}}}$ can be directly fed into the lightweight networks. Then the feature representation $h_{\mathcal{S}}$ of student PCP model is obtained by $\textbf{Net}({\bf{\hat{\emph{B}}}})$, where \textbf{Net}() denotes the lightweight network.

With $h_{\mathcal{S}}$, the binary classifier $\phi(h_{\mathcal{S}})$ is the same as $\phi(h_{\mathcal{T}})$ modeled by a Global Average Pooling (GAP) layer, 2 fully-connected layers, and generates the student logits $z_{\mathcal{S}}$ which is classified by a \emph{softmax} layer for crossing or not-crossing determination. 

\subsection{Virtual-to-Real Distillation}

With the modeling of teacher PCP model $\mathcal{T}$ and student PCP model $\mathcal{S}$, the virtual-to-real distillation is described in this subsection. Assume the teacher PCP model is trained offline with the synthetic data, the distillation process is achieved by inputing the real data into the trained $\mathcal{T}$ and $\mathcal{S}$ to be trained simultaneously. Notably, in order to adsorb the representation ability for $\mathcal{T}$, we need to obtain the same information configuration as the one of synthetic data. Therefore, before distillation, we first generate the optical flow images on the real video data. Then, the full video frames of real data, optical flow images, the location and local visual context of pedestrians of real data are fed into $\mathcal{T}$ for generating the logits $z_{\mathcal{T}}$ of teacher model. Only the locations of pedestrians are inputted into $\mathcal{S}$ and generates logits $z_{\mathcal{S}}$. 

As aforementioned in Eq. \ref{eq1}, $\mathcal{R}(\zeta(z_\mathcal{T},T),  \zeta(z_\mathcal{S},T))$ and $\mathcal{H}(h_\mathcal{T},h_\mathcal{S})$ bridge the distillation process between $\mathcal{T}$ and $\mathcal{S}$.  In this work, we adopt the Kullback-Leibler divergence (KLD) and mean square log loss to define the response distillation function $\mathcal{R}(.)$ and the feature distillation function $\mathcal{H}(.)$, and defined as:
\begin{equation}
\begin{array}{ll} 
\mathcal{R}(\zeta_{\mathcal{T}}, \zeta_{\mathcal{S}})=\sum_i(\zeta_{\mathcal{T}}(i)\log(\zeta_{\mathcal{T}}(i))-\zeta_{\mathcal{S}}(i)\log(\zeta_{\mathcal{S}}(i)),\\\\
\mathcal{H}(h_\mathcal{T},h_\mathcal{S})=\sum_i(\log(1+h_\mathcal{T}(i))-\log(1+h_\mathcal{S}(i)))^{2},
\end{array}
\end{equation}
where $i$ is the index of the value in the logits $z_\mathcal{S}$ or $z_\mathcal{T}$. $\zeta_{\mathcal{T}}$=$\zeta(z_\mathcal{T},T))$, $\zeta_{\mathcal{S}}$=$\zeta(z_\mathcal{S},T))$, and $\zeta(.,T))$ denotes the \emph{softmax} function with the hyper-parameter temperature $T$ which is set as 2 in this work. 

With these modeling for the modules in VR-PCP, the distillation process can be conducted by minimizing the function defined in Eq. \ref{eq1}. In the following, we will evaluate the proposed method with exhaustive experiments.

\section{EXPERIMENTS}
\subsection{Dataset: Virtual-PedCross-4667}

 \begin{figure}[!t]
  \centering
  \includegraphics[width=0.9\hsize]{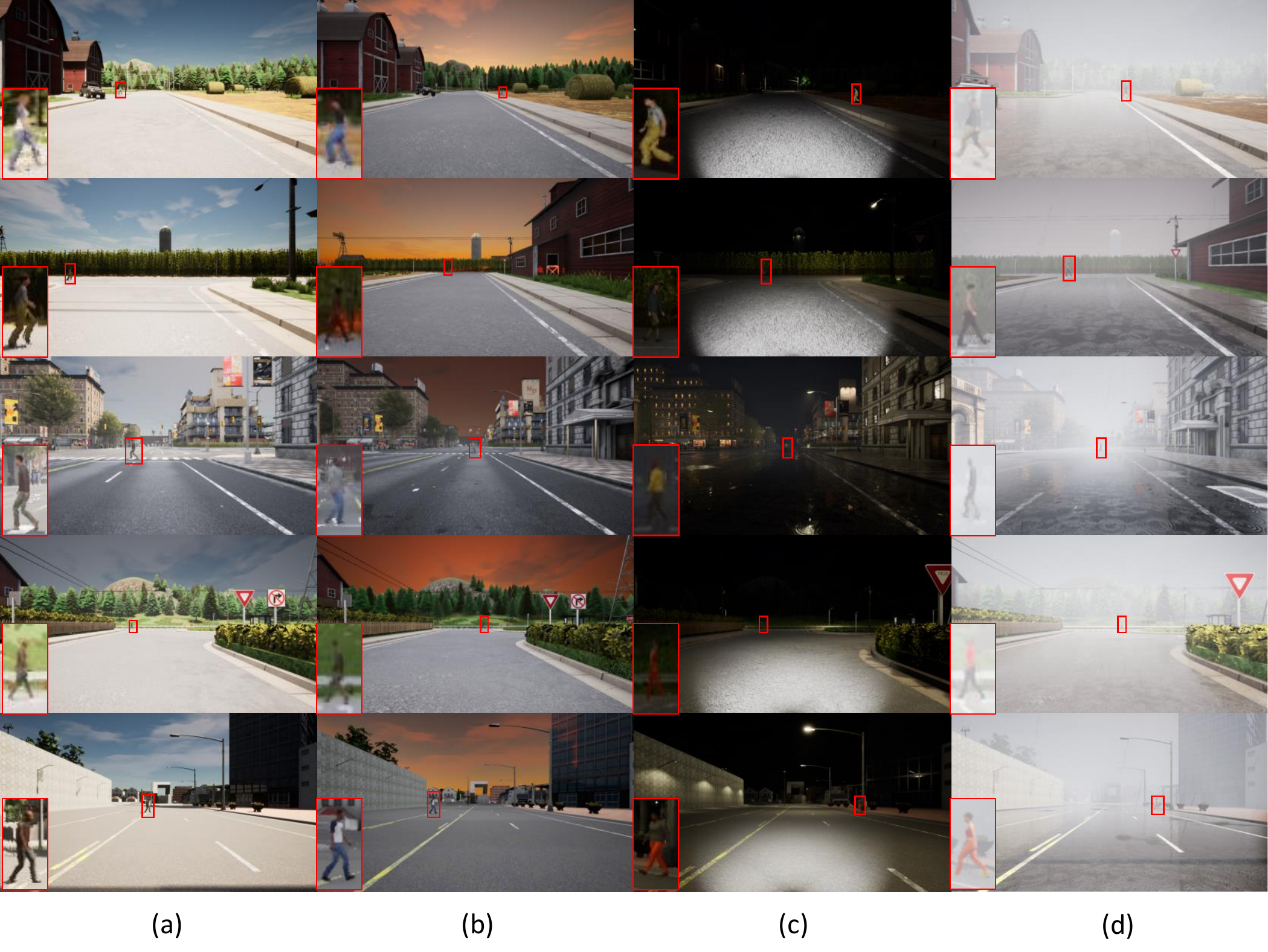}
  \caption{Some pedestrian crossing examples in Virtual-Pedcross-4667 dataset. The crossing pedestrians are marked by red bounding boxes. The same column presents the samples under the same scene with different lighting and weather conditions, where (a), (b), (c), and (d) represent sunny, evening, nighttime and rainy conditions.}
  \label{fig4}
\end{figure}

In this work, we use CARLA simulator to generate a large amount of virtual driving videos with pedestrian crossing behavior. Specially, the scene maps, weather and light condition, pedestrian age and gender are considered in the data generation. Following JAAD \cite{DBLP:conf/iccvw/RasouliKT17} and PIE \cite{DBLP:conf/iccv/RasouliKKT19} datasets, the forward dashcam videos are collected, and 4667 video sequences are collected (called Virtual-PedCross-4667), consisting of 2862 pedestrian crossing sequences and 1804 not-crossing sequences. Totally, 745k video frames with the resolution of 1280$\times$720 are saved. Some typical examples of pedestrian crossing in the Virtual-Pedcross-4667 dataset are shown in Fig. \ref{fig4}. In the pedestrian crossing sequence, there is one pedestrian in each sequence exhibiting crossing behavior, while for the pedestrian not-crossing sequence, each sequence will randomly appear 1 to 3 pedestrians who do not cross. This setting is helpful for treating an entire video sequence as a positive sample or a negative sample. Each pedestrian crossing sequence contains 200 video frames, while the non-crossing sequence contains 100 video frames. Each frame will be automatically marked with five attributes: occasions, weather, gender, age, and the bounding box coordinate of pedestrian. 
  \begin{table}[htpb]
	\centering
	\caption{Comparison of Virtual-PedCross-4667 with other existing pedestrian crossing datasets. S./R.: synthetic or real data.}
	\resizebox{0.5\textwidth}{!}{
		\begin{tabular}{c|c|c|c|c}
    	\hline\hline
    	dataset & \#seqs. & \#frames  & \#peds & S./R. \\
    	\hline
		JAAD \cite{DBLP:conf/iccvw/RasouliKT17} & 346 & 75K & 686 & R   \\
		PIE  \cite{DBLP:conf/iccv/RasouliKKT19} & 55 & 293K & 1800 & R   \\
		CP2A \cite{DBLP:journals/corr/abs-2107-08031}  & 77500 & 2.325M & - & S \\
		Virtual-PedCross-4667 & 4667 & 745K & 5835 & S  \\
    	\hline\hline
    	\end{tabular}}
	\label{tab1}
\end{table}

 \begin{figure}[!t]
  \centering
  \includegraphics[width=\hsize]{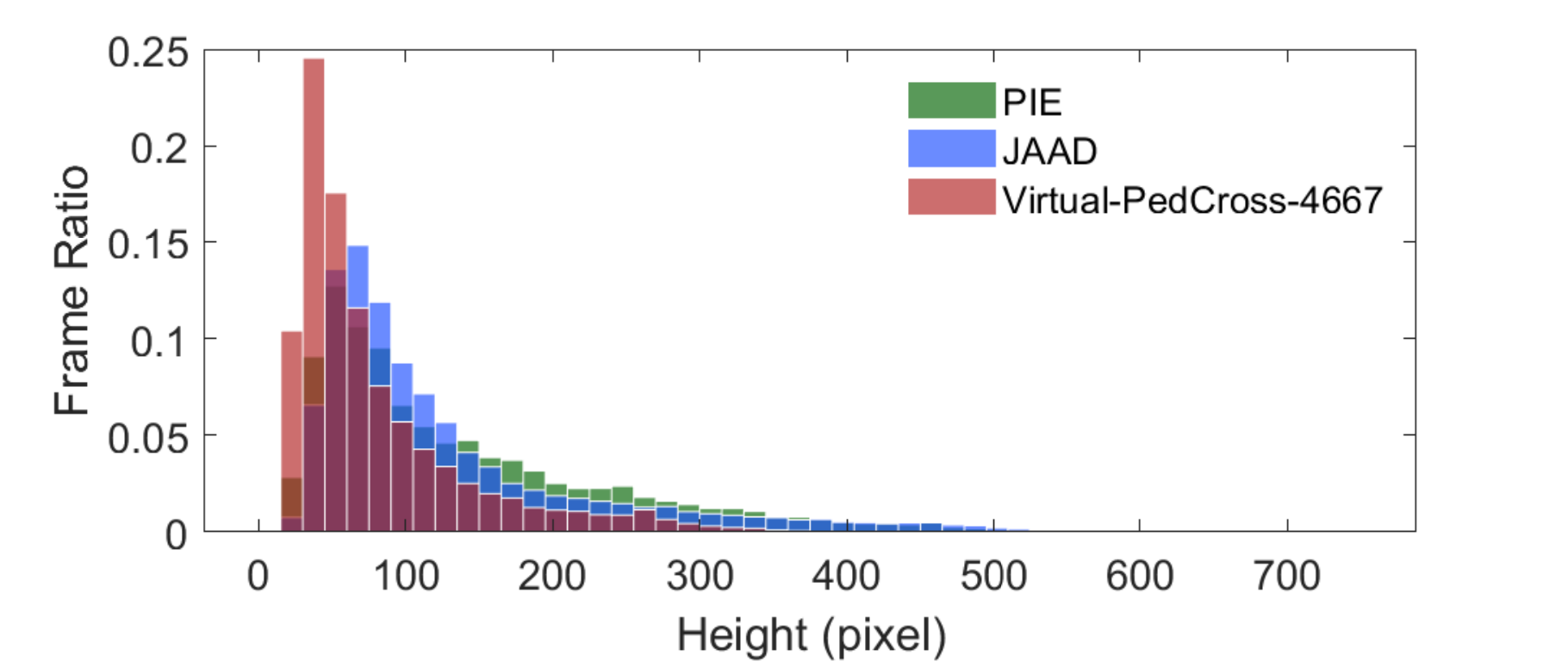}
  \caption{Height (in pixels) distribution comparison between Virtual-PedCross-4667, PIE and JAAD datasets.}
  \label{fig5}
\end{figure}

We compare our dataset statistics with JAAD, PIE, and CP2A \cite{DBLP:journals/corr/abs-2107-08031} (recently reported) in terms of sequence numbers, frame numbers, and pedestrian counts, as shown in Table \ref{tab1}. Furthermore, since the pedestrian scale has a large impact on the prediction accuracy, we compare the pedestrian scale statistics of our dataset, JAAD and PIE in Fig. \ref{fig5}. From this figure, we can see that our dataset covers more samples with small scale of pedestrians, which is useful for crossing behavior prediction in early time.

\subsection{Implementation Details}
We train the model $\mathcal{T}$ with Virtual-PedCross-4667, and train the model $\mathcal{S}$ with PIE and JAAD datasets. All video frames are scaled to 224$\times$224, so the input dimension is set as [batchsize, $N$, 224, 224, 3]. The non-vision information dimension of pedestrian bounding boxes is [batch size, $N$, 4]. The batch size in this work is set as 2. The number of observation frames $N$ is set as 16 with 0.5 seconds (30fps). Time to Crossing (TTC) is set as 1-2 seconds (30-60 frames). During testing,  although $\mathcal{T}$ requires rich pedestrian crossing information, we only evaluate the performance of $\mathcal{S}$, which will not influence the execution efficiency of $\mathcal{S}$.

For training $\mathcal{T}$,  we train the model $\mathcal{T}$ with epoch of 20 with learning rate of $5\times10^{-5}$. A dropout of 0.5 is added during training. After obtaining the trained model $\mathcal{T}$, we conduct the distillation process and train the model $\mathcal{S}$ by using Adam optimizer with a learning rate of $5\times10^{-5}$ and epoch of 60 and 120 for PIE and JAAD datasets, respectively. The training and testing settings are the same as other updated works \cite{DBLP:conf/wacv/KotserubaRT21,DBLP:conf/fgr/GesnouinPSM21,DBLP:journals/corr/abs-2104-05485}.

\textbf{Metrics:} Following the updated works, we take \emph{accuracy} (Acc), \emph{F1 score} (F1), \emph{precision} (Pre) and \emph{recall} (Rec), and the area under curve (Auc) metrics to evaluate the performance. These metrics prefer a larger value.

\subsection{Ablation Studies}
In this work, we input four kinds of information into the teacher PCP model. Which combination is the best for pedestrian crossing prediction and the distillation process? We exhaustively evaluate the performance difference with different information combination in the distillation process. In the comparison, we also switch the lightweight student PCP model, i.e., Transformer (Trans.), ResNet18 (R.Net), MobileNet (M.Net) and ShuffleNet (S.Net). The performance comparison results are demonstrated in Table. \ref{tab2}. From the results, we can see that the performance increases with more information, and the lightweight model Transformer generates the best performance compared with other student PCP models. Notably, we also present the results for the student PCP model without distillation process (BB w/o distill.).

Because the number of negative samples is more than the positive samples in JAAD dataset, almost all the recall value is larger than the precision value for the methods. Actually, the sample imbalance issue is important in the distillation process. From the result of ``BB w/o distill", we can see than without the distillation process, the gap between recall and precision values is larger than the ones after distillation. Therefore, with the help of teacher PCP model, we can restrict the sample imbalance issue better.
\begin{table}[!t]\scriptsize
\centering
\caption{Performance comparison of student PCP, w.r.t., different input Informations (Info. ) and different lightweight Student Networks (Net.).The local context region, bounding boxes, the local motion region and global context region are abbreviated as LC, BB, LM, GC, respectively.}
	\begin{tabular}{c|c|c|c|c|c|c}
	\hline\hline
	\multirow{2}{*}{Net.}&\multirow{2}{*}{Info.} & \multicolumn{5}{c}{JAAD\_all} \\	
	\cmidrule{3-7} &&Acc & Auc & F1 & Pre & Rec  \\
	\cline{1-7}
\multirow{5}{*}{R.Net}&BB w/o distill.& 0.75 & 0.68 & 0.52 & 0.39 & 0.71 \\
\cline{2-7}
&BB & 0.80 & 0.68 & 0.50 & 0.51 & 0.50 \\
&BB\&LC & 0.81 & 0.69 & 0.58 & 0.60 & 0.57 \\
&BB\&LC\&GC & 0.82 & 0.74 & 0.65 & 0.61 & 0.70 \\
&BB\&LC\&GC\&LM& 0.82 & 0.77 & 0.56 & 0.58 & 0.68 \\
\hline
\multirow{5}{*}{M.Net}&BB w/o distill.& 0.71 & 0.62 & 0.36 & 0.29 & 0.49 \\
\cline{2-7}
&BB & 0.80 & 0.74 & 0.52 & 0.47 & 0.60 \\
&BB\&LC & 0.80 & 0.74 & 0.63 & 0.55 & 0.66 \\
&BB\&LC\&GC & 0.82 & 0.75& 0.63 & 0.57 & 0.64 \\
&BB\&LC\&GC\&LM & 0.82 & 0.74 & 0.64 & 0.58 & 0.63 \\
\hline
\multirow{5}{*}{S.Net}&all w/o distill.& 0.72 & 0.73 & 0.45 & 0.28 & 0.49 \\
\cline{2-7}
&BB & 0.83 & 0.69 & 0.51 & 0.58 & 0.45 \\
&BB\&LC & 0.82 & 0.68 & 0.51 & 0.65 & 0.41 \\
&BB\&LC\&GC & 0.85 & 0.73 & 0.54 & 0.52 & 0.56 \\
&BB\&LC\&GC\&LM& 0.85 & 0.74 & 0.56 & 0.56 & 0.56 \\
\hline
\multirow{5}{*}{Trans.}&BB w/o distill.& 0.75 & 0.77 & 0.52 & 0.49 & 0.69 \\
\cline{2-7}
&BB & 0.85 & 0.71 & 0.62 & 0.66 & 0.58 \\
&BB\&LC & 0.85 & 0.70 & 0.62 & 0.66 & 0.69 \\
&BB\&LC\&GC & 0.84 & 0.78 & 0.77 & 0.68 & 0.79 \\
&BB\&LC\&GC\&LM& 0.86 & 0.81 & 0.77 & 0.74 & 0.81 \\
\hline\hline
   \end{tabular}
\label{tab2}
\end{table}

\begin{table*}[htpb]\footnotesize
\centering
\caption{Performance comparison of the baselines and the state-of-the-art on JAAD$_{all}$, JAAD$_{behavior}$, and PIE datasets.}
\begin{tabular}{ccccccc|rrrrr|ccccc}
\hline\hline
\multicolumn{ 2}{c}{Model} &   \multicolumn{5}{c|}{JAAD$_{all}$} &  \multicolumn{5}{c|}{JAAD$_{behavior}$} & \multicolumn{5}{c}{PIE} \\
\cline{3-17}
\multicolumn{ 2}{c}{} & Acc & Auc &  F1 &  Pre &  Rec & Acc & Auc & F1 & Pre & Rec & Acc & Auc & F1 & Pre & Rec \\
\hline
\multicolumn{ 2}{c}{ATGC} \cite{DBLP:conf/iccvw/RasouliKT17} &       0.67 &       0.62 &       0.76 &       0.72 &      0.80  &       0.48 &       0.41 &       0.62 &       0.58 &       0.66 &       0.59 &       0.55 &       0.39 &       0.33 &       0.47 \\
\hline
\multicolumn{ 2}{c}{SPI-Net} \cite{DBLP:journals/algorithms/GesnouinPBSM20}&       0.81 &       0.72 &       0.52 &       0.48 &       0.58 &       0.58 &       0.55 &       0.66 &       0.67 &       0.65 &       0.66 &       0.54 &      0.30  &       0.35 &       0.27 \\

\multicolumn{2}{c}{SingleRNN} \cite{DBLP:conf/ivs/KotserubaRT20}&       0.78 &       0.75 &       0.54 &       0.44 &      0.70  &       0.51 &       0.48 &       0.61 &       0.63 &       0.59 &       0.81 &       0.75 &       0.64 &       0.67 &       0.61 \\
\multicolumn{ 2}{c}{MultiRNN} \cite{DBLP:conf/cvpr/BhattacharyyaFS18}&       0.79 &       0.79 &       0.58 &       0.45 &       0.79 &       0.61 &      0.50  &       0.74 &       0.64 &       0.86 &       0.83 &      0.80  &       0.71 &       0.69 &       0.73 \\

\multicolumn{ 2}{c}{SFRNN} \cite{DBLP:conf/bmvc/RasouliKT19}&       0.84 &       \textbf{0.84} &       0.65 &       0.54 &       \textbf{0.84} &       0.51 &       0.45 &       0.63 &       0.61 &       0.64 &       0.82 &       0.79 &       0.69 &       0.67 &      0.70  \\
\hline
\multicolumn{ 2}{c}{I3D} \cite{DBLP:conf/cvpr/CarreiraZ17}&       0.81 &       0.74 &       0.63 &       0.66 &       0.61 &       0.62 &       0.56 &       0.73 &       0.68 &       0.79 &      0.80  &       0.73 &       0.62 &       0.67 &       0.58 \\

\multicolumn{ 2}{c}{PCPA} \cite{DBLP:conf/wacv/KotserubaRT21}&       0.83 &       0.83 &       0.64 &0.55 &   - &       0.56 &      0.56  &       0.68 &  - & - &       0.87 &       0.86 &       0.77 &   -   &-\\
\multicolumn{ 2}{c}{TrousSPI-Net} \cite{DBLP:conf/fgr/GesnouinPSM21} &       0.85 &       0.73 &       0.56 &0.57 &  0.55&\textbf{0.64} &       0.56 &      0.76  &       0.66 &  0.91 & 0.88 &       \textbf{0.88} &       \textbf{0.80} &       0.73  &\textbf{0.89}\\
\multicolumn{ 2}{c}{FF-ST-Attention} \cite{DBLP:journals/corr/abs-2104-05485}&       0.83 &       0.82 &       0.63 &0.51 &   0.81 &       0.62 &      0.54  &       0.74 &  0.65& 0.85&       - &       - &       - &   -   &-\\
\hline
\multicolumn{2}{c}{Ours(R.Net)} &       0.82 &       0.77 &       0.56 &       0.58 &      0.68  &        0.60 &       0.57 &       \textbf{0.77} &       0.67 &       0.91 &       0.87 &       \textbf{0.88} &       \textbf{0.80} &       0.73 &      0.86 \\

\multicolumn{ 2}{c}{Ours(M.Net)} &       0.82 &       0.74 &       0.64 &       0.58 &       0.63 &       0.61 &        0.50 &       0.75 &       0.71 &        0.80 &       0.84 &       0.83 &       0.74 &       0.68 &        0.80 \\

\multicolumn{2}{c}{Ours(S.Net)} &       0.85 &       0.74 &       0.56 &       0.56 &      0.56  &        0.60 &       0.49 &       0.75 &        0.60 &       \textbf{0.99} &       0.77 &       0.64 &       0.46 &       0.68 &       0.35 \\
\multicolumn{2}{c}{Ours(Trans.)} &       \textbf{0.86} &       0.81 &       \textbf{0.77} &      \textbf{ 0.74} &      0.81 &    \textbf{0.64}&       \textbf{0.66}&            0.76&           0.70&         0.89& \textbf{0.89} &       \textbf{0.88} &       0.67 &       \textbf{0.74} &       0.84 \\
\hline\hline
\end{tabular}
\label{tab3}
\end{table*}
  
\subsection{Comparison with State-of-The-Art}
In this work, we compare the proposed method with nine state-of-the-art approaches on JAAD dataset and PIE dataset. The methods are listed in Table \ref{tab3}, where the ATGC is modeled by the traditional AlexNet. SPI-Net \cite{DBLP:journals/algorithms/GesnouinPBSM20}, SingleRNN \cite{DBLP:conf/ivs/KotserubaRT20}, MultiRNN \cite{DBLP:conf/cvpr/BhattacharyyaFS18}, and SFRNN  \cite{DBLP:conf/bmvc/RasouliKT19} take the same Gated Recurrent Unit (GRU) to model the temporal feature of observation of pedestrian movement. Actually, most of the competitors in this work consider multiple information for feature encoding of sequence observation, while our work (student PCP model) with the virtual-to-real distillation only take the bounding box into consideration. Meanwhile, most of the works take the same setting with 0.5s observation and 1-2s of time-to-crossing (TTC). The results in Table. \ref{tab3} are reported from their works.

 \begin{figure}[!t]
  \centering
  \includegraphics[width=\hsize]{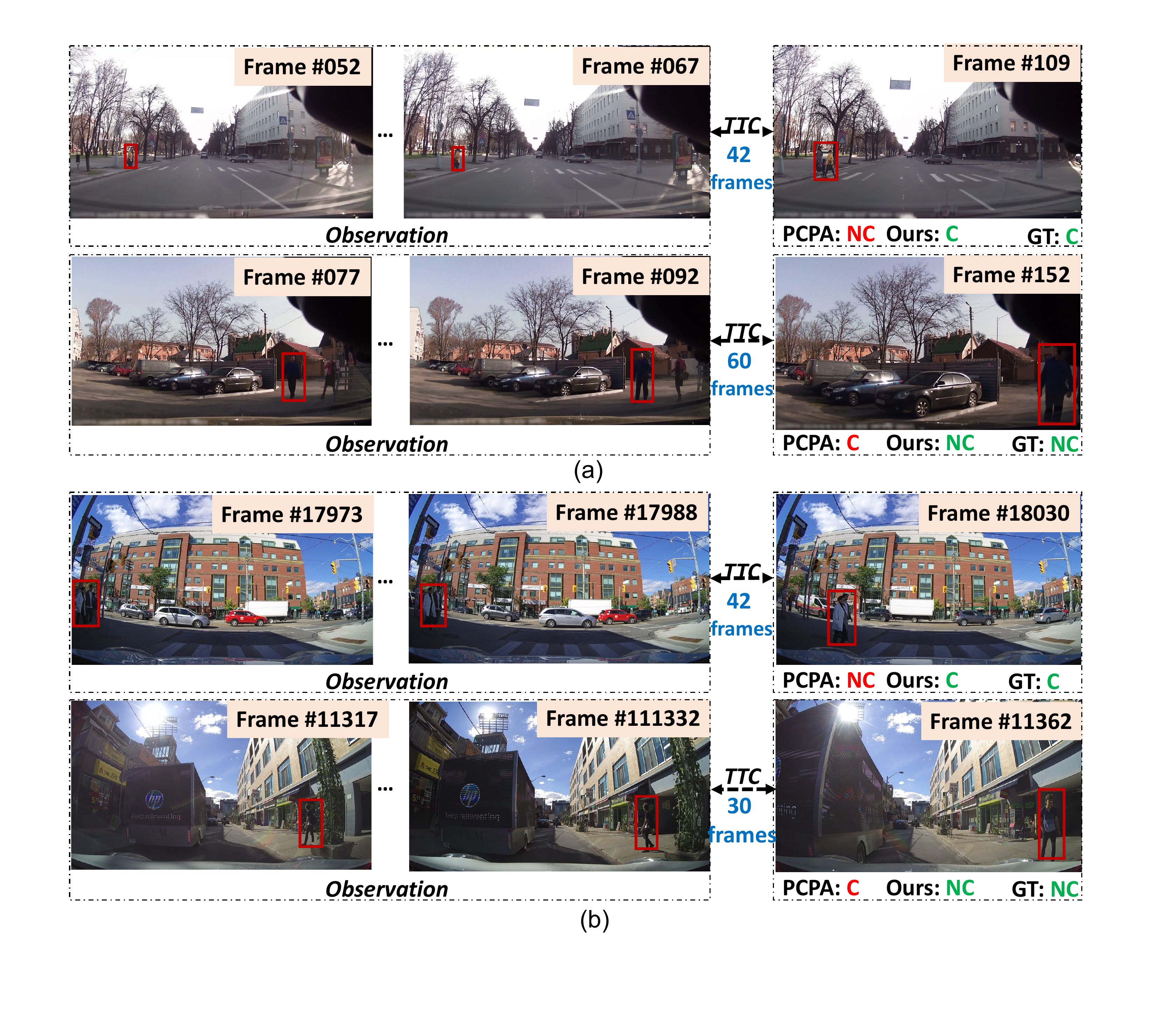}
  \caption{Some frameshots of predicted results of crossing (C) or not crossing (NC) in (a) JAAD dataset and (b) PIE dataset by PCPA \cite{DBLP:conf/wacv/KotserubaRT21} and our method. GT represents the ground-truth. The pedestrians are marked by red boxes.}
  \label{fig6}
\end{figure}
From Table. \ref{tab3}, we can see that our methods, especially for the ``\emph{Ours (Trans.)}“ and ``\emph{Ours (R. Net)}“ generate the comparative performance on PIE dataset, and ``\emph{Ours (Trans.)}“  is best for JAAD dataset. TrousSPI-Net demonstrates good performance on PIE dataset, which fuses bounding boxes, vehicle speed, pedestrian pose together and own more parameters than our work. Therefore, we can conclude that the distillation process from virtual to real dataset is useful for pedestrian crossing prediction.  

In Fig. \ref{fig6}, we also demonstrate some frameshots with the comparison between  PCPA \cite{DBLP:conf/wacv/KotserubaRT21} with weight size of 29.72M  and our method (Ours (Trans.)) with weight size of 4.77M. From the demonstrated frames, the crossing pedestrians show small scale in the beginning and the not-crossing pedestrians have the same moving direction with crossing in the observation, which harden the prediction, and our method shows promising results with rather less computation cost.

\section{Conclusions}

In this paper, we propose a deep virtual-to-real distillation framework for pedestrian crossing prediction in driving scenarios. A large-scale virtual dataset, called Virtual-Pedcross-4667 with 745k frames were constructed with careful consideration for light and weather conditions. Through the distillation process, we can simply the architecture of student pedestrian crossing prediction model and generate promising prediction performance. Based on the comparison with many state-of-the-art methods, the superiority of the proposed method is validated by exhaustive experiments. In the future, we will explore more advanced knowledge distillation frameworks and better teacher and student models. In addition, we will collect more crossing scenarios on real-world roads to enhance the reliability of the results.

\scriptsize{
\bibliographystyle{IEEEtran}
\bibliography{ref}

\begin{thebibliography}{10}
\providecommand{\url}[1]{#1}
\csname url@samestyle\endcsname
\providecommand{\newblock}{\relax}
\providecommand{\bibinfo}[2]{#2}
\providecommand{\BIBentrySTDinterwordspacing}{\spaceskip=0pt\relax}
\providecommand{\BIBentryALTinterwordstretchfactor}{4}
\providecommand{\BIBentryALTinterwordspacing}{\spaceskip=\fontdimen2\font plus
\BIBentryALTinterwordstretchfactor\fontdimen3\font minus
  \fontdimen4\font\relax}
\providecommand{\BIBforeignlanguage}[2]{{%
\expandafter\ifx\csname l@#1\endcsname\relax
\typeout{** WARNING: IEEEtran.bst: No hyphenation pattern has been}%
\typeout{** loaded for the language `#1'. Using the pattern for}%
\typeout{** the default language instead.}%
\else
\language=\csname l@#1\endcsname
\fi
#2}}
\providecommand{\BIBdecl}{\relax}
\BIBdecl

\bibitem{trafficdeath}
W.~H. Organization, ``Global health observatory: Number of road traffic
  deaths,''
  \url{https://www.who.int/gho/road-safety/mortality/traffic-deaths-number/en/}.

\bibitem{DBLP:conf/iros/WangP20}
Z.~Wang and N.~Papanikolopoulos, ``Estimating pedestrian crossing states based
  on single 2d body pose,'' in \emph{IROS}, 2020, pp. 2205--2210.

\bibitem{DBLP:conf/itsc/CadenaYQW19}
P.~R.~G. Cadena, M.~Yang, Y.~Qian, and C.~Wang, ``Pedestrian graph: Pedestrian
  crossing prediction based on 2d pose estimation and graph convolutional
  networks,'' in \emph{ITSC}, 2019, pp. 2000--2005.

\bibitem{DBLP:journals/corr/abs-2107-08031}
L.~Achaji, J.~Moreau, T.~Fouqueray, F.~Aioun, and F.~Charpillet, ``Is attention
  to bounding boxes all you need for pedestrian action prediction?''
  \emph{CoRR}, vol. abs/2107.08031, 2021.

\bibitem{DBLP:conf/cvpr/CarreiraZ17}
J.~Carreira and A.~Zisserman, ``Quo vadis, action recognition? {A} new model
  and the kinetics dataset,'' in \emph{CVPR}, 2017, pp. 4724--4733.

\bibitem{DBLP:conf/bmvc/RasouliKT19}
A.~Rasouli, I.~Kotseruba, and J.~K. Tsotsos, ``Pedestrian action anticipation
  using contextual feature fusion in stacked rnns,'' in \emph{BMVC}, 2019, p.
  171.

\bibitem{DBLP:journals/corr/abs-2104-05485}
D.~Yang, H.~Zhang, E.~Yurtsever, K.~A. Redmill, and
  {\"{U}}.~{\"{O}}zg{\"{u}}ner, ``Predicting pedestrian crossing intention with
  feature fusion and spatio-temporal attention,'' \emph{CoRR}, vol.
  abs/2104.05485, 2021.

\bibitem{DBLP:conf/itsc/XueLCZHZ19}
P.~Xue, J.~Liu, S.~Chen, Z.~Zhou, Y.~Huo, and N.~Zheng, ``Crossing-road
  pedestrian trajectory prediction via encoder-decoder {LSTM},'' in
  \emph{ITSC}, 2019, pp. 2027--2033.

\bibitem{DBLP:conf/ivs/LorenzoPWSLS20}
J.~Lorenzo, I.~Parra, F.~Wirth, C.~Stiller, D.~F. Llorca, and M.~{\'{A}}.
  Sotelo, ``Rnn-based pedestrian crossing prediction using activity and
  pose-related features,'' in \emph{IV}, 2020, pp. 1801--1806.

\bibitem{DBLP:journals/sensors/0002AIBHLS21}
J.~Lorenzo \emph{et~al.}, ``Capformer: Pedestrian crossing action prediction
  using transformer,'' \emph{Sensors}, vol.~21, no.~17, p. 5694, 2021.

\bibitem{DBLP:conf/wacv/KotserubaRT21}
I.~Kotseruba, A.~Rasouli, and J.~K. Tsotsos, ``Benchmark for evaluating
  pedestrian action prediction,'' in \emph{WACV}, 2021, pp. 1257--1267.

\bibitem{DBLP:conf/iccvw/RasouliKT17}
A.~Rasouli, I.~Kotseruba, and J.~K. Tsotsos, ``Are they going to cross? {A}
  benchmark dataset and baseline for pedestrian crosswalk behavior,'' in
  \emph{ICCVW}, 2017, pp. 206--213.

\bibitem{Dosovitskiy17}
A.~Dosovitskiy, G.~Ros, F.~Codevilla, A.~Lopez, and V.~Koltun, ``{CARLA}: {An}
  open urban driving simulator,'' in \emph{Proceedings of the 1st Annual
  Conference on Robot Learning}, 2017, pp. 1--16.

\bibitem{DBLP:conf/iccv/RasouliKKT19}
A.~Rasouli, I.~Kotseruba, T.~Kunic, and J.~K. Tsotsos, ``{PIE:} {A} large-scale
  dataset and models for pedestrian intention estimation and trajectory
  prediction,'' in \emph{ICCV}, 2019, pp. 6261--6270.

\bibitem{DBLP:journals/pami/KongTF20}
Y.~Kong \emph{et~al.}, ``Adversarial action prediction networks,'' \emph{{IEEE}
  Trans. Pattern Anal. Mach. Intell.}, vol.~42, no.~3, pp. 539--553, 2020.

\bibitem{DBLP:journals/ijcv/WuWHLL21}
X.~Wu, R.~Wang, J.~Hou, H.~Lin, and J.~Luo, ``Spatial-temporal relation
  reasoning for action prediction in videos,'' \emph{Int. J. Comput. Vis.},
  vol. 129, no.~5, pp. 1484--1505, 2021.

\bibitem{DBLP:journals/tcsv/ChenLSZ21}
L.~Chen, J.~Lu, Z.~Song, and J.~Zhou, ``Recurrent semantic preserving
  generation for action prediction,'' \emph{{IEEE} Trans. Circuits Syst. Video
  Technol.}, vol.~31, no.~1, pp. 231--245, 2021.

\bibitem{DBLP:journals/tip/RoyF21}
D.~Roy and B.~Fernando, ``Action anticipation using pairwise human-object
  interactions and transformers,'' \emph{{IEEE} Trans. Image Process.},
  vol.~30, pp. 8116--8129, 2021.

\bibitem{DBLP:conf/eccv/LiuTLR20}
M.~Liu, S.~Tang, Y.~Li, and J.~M. Rehg, ``Forecasting human-object interaction:
  Joint prediction of motor attention and actions in first person video,'' in
  \emph{ECCV}, vol. 12346, 2020, pp. 704--721.

\bibitem{DBLP:conf/eccv/LanCS14}
T.~Lan, T.~Chen, and S.~Savarese, ``A hierarchical representation for future
  action prediction,'' in \emph{ECCV}, vol. 8691, 2014, pp. 689--704.

\bibitem{DBLP:conf/cvpr/KongTF17}
Y.~Kong, Z.~Tao, and Y.~Fu, ``Deep sequential context networks for action
  prediction,'' in \emph{CVPR}, 2017, pp. 3662--3670.

\bibitem{DBLP:journals/tcsv/HuangLLLL20}
J.~Huang, N.~Li, T.~H. Li, S.~Liu, and G.~Li, ``Spatial-temporal context-aware
  online action detection and prediction,'' \emph{{IEEE} Trans. Circuits Syst.
  Video Technol.}, vol.~30, no.~8, pp. 2650--2662, 2020.

\bibitem{DBLP:conf/ijcai/ZhangMYLJR21}
T.~Zhang, W.~Min, J.~Yang, T.~Liu, S.~Jiang, and Y.~Rui, ``What if we could not
  see? counterfactual analysis for egocentric action anticipation,'' in
  \emph{IJCAI}, 2021, pp. 1316--1322.

\bibitem{DBLP:conf/itsc/WuWZXW20}
H.~Wu, L.~Wang, S.~Zheng, Q.~Xu, and J.~Wang, ``Crossing-road pedestrian
  trajectory prediction based on intention and behavior identification,'' in
  \emph{ITSC}, 2020, pp. 1--6.

\bibitem{DBLP:conf/ijcai/YaoAJV021}
Y.~Yao, E.~M. Atkins, M.~Johnson{-}Roberson, R.~Vasudevan, and X.~Du,
  ``Coupling intent and action for pedestrian crossing behavior prediction,''
  in \emph{IJCAI}, 2021, pp. 1238--1244.

\bibitem{DBLP:conf/iros/SchneemannH16}
F.~Schneemann and P.~Heinemann, ``Context-based detection of pedestrian
  crossing intention for autonomous driving in urban environments,'' in
  \emph{IROS}, 2016, pp. 2243--2248.

\bibitem{DBLP:conf/aaai/CaiLHZ19}
Y.~Cai, H.~Li, J.~Hu, and W.~Zheng, ``Action knowledge transfer for action
  prediction with partial videos,'' in \emph{AAAI}, 2019, pp. 8118--8125.

\bibitem{DBLP:conf/icpr/CamporeseCFFB20}
G.~Camporese, P.~Coscia, A.~Furnari, G.~M. Farinella, and L.~Ballan,
  ``Knowledge distillation for action anticipation via label smoothing,'' in
  \emph{ICPR}, 2020, pp. 3312--3319.

\bibitem{DBLP:conf/cvpr/WangHLZZ19}
X.~Wang, J.~Hu, J.~Lai, J.~Zhang, and W.~Zheng, ``Progressive teacher-student
  learning for early action prediction,'' in \emph{CVPR}, 2019, pp. 3556--3565.

\bibitem{DBLP:conf/cvpr/IlgMSKDB17}
E.~Ilg, N.~Mayer, T.~Saikia, M.~Keuper, A.~Dosovitskiy, and T.~Brox, ``Flownet
  2.0: Evolution of optical flow estimation with deep networks,'' in
  \emph{CVPR}, 2017, pp. 1647--1655.

\bibitem{DBLP:conf/iccv/TranBFTP15}
D.~Tran, L.~D. Bourdev, R.~Fergus, L.~Torresani, and M.~Paluri, ``Learning
  spatiotemporal features with 3d convolutional networks,'' in \emph{ICCV},
  2015, pp. 4489--4497.

\bibitem{DBLP:conf/nips/VaswaniSPUJGKP17}
A.~Vaswani, N.~Shazeer, N.~Parmar, J.~Uszkoreit, L.~Jones, A.~N. Gomez,
  L.~Kaiser, and I.~Polosukhin, ``Attention is all you need,'' in
  \emph{NeurIPS}, 2017, pp. 5998--6008.

\bibitem{DBLP:conf/cvpr/HeZRS16}
K.~He, X.~Zhang, S.~Ren, and J.~Sun, ``Deep residual learning for image
  recognition,'' in \emph{CVPR}, 2016, pp. 770--778.

\bibitem{DBLP:journals/corr/HowardZCKWWAA17}
A.~G. Howard, M.~Zhu, B.~Chen, D.~Kalenichenko, W.~Wang, T.~Weyand,
  M.~Andreetto, and H.~Adam, ``Mobilenets: Efficient convolutional neural
  networks for mobile vision applications,'' \emph{CoRR}, vol. abs/1704.04861,
  2017.

\bibitem{DBLP:conf/cvpr/ZhangZLS18}
X.~Zhang, X.~Zhou, M.~Lin, and J.~Sun, ``Shufflenet: An extremely efficient
  convolutional neural network for mobile devices,'' in \emph{CVPR}, 2018, pp.
  6848--6856.

\bibitem{DBLP:conf/fgr/GesnouinPSM21}
J.~Gesnouin, S.~Pechberti, B.~Stanciulcscu, and F.~Moutarde, ``Trouspi-net:
  Spatio-temporal attention on parallel atrous convolutions and u-grus for
  skeletal pedestrian crossing prediction,'' in \emph{FG}, 2021, pp. 1--7.

\bibitem{DBLP:journals/algorithms/GesnouinPBSM20}
J.~Gesnouin, S.~Pechberti, G.~Bresson, B.~Stanciulescu, and F.~Moutarde,
  ``Predicting intentions of pedestrians from 2d skeletal pose sequences with a
  representation-focused multi-branch deep learning network,''
  \emph{Algorithms}, vol.~13, no.~12, p. 331, 2020.

\bibitem{DBLP:conf/ivs/KotserubaRT20}
I.~Kotseruba, A.~Rasouli, and J.~K. Tsotsos, ``Do they want to cross?
  understanding pedestrian intention for behavior prediction,'' in \emph{IV},
  2020, pp. 1688--1693.

\bibitem{DBLP:conf/cvpr/BhattacharyyaFS18}
A.~Bhattacharyya, M.~Fritz, and B.~Schiele, ``Long-term on-board prediction of
  people in traffic scenes under uncertainty,'' in \emph{CVPR}, 2018, pp.
  4194--4202.

\end{thebibliography}
}

\end{document}